# NUGGET DISCOVERY WITH A MULTI-OBJECTIVE CULTURAL ALGORITHM

Sujatha Srinivasan[1] and Sivakumar Ramakrishnan[2]

[1,2]Department of Computer Science, AVVM Sri Pushpam College, Poondi, Tamil Nadu
India

ashoksuja03@yahoo.co.in
rskumar.avvmspc@gmail.com

## ABSTRACT

*Partial classification popularly known as nugget discovery comes under descriptive knowledge discovery. It involves mining rules for a target class of interest. Classification "If-Then" rules are the most sought out by decision makers since they are the most comprehensible form of knowledge mined by data mining techniques. The rules have certain properties namely the rule metrics which are used to evaluate them. Mining rules with user specified properties can be considered as a multi-objective optimization problem since the rules have to satisfy more than one property to be used by the user. Cultural algorithm (CA) with its knowledge sources have been used in solving many optimization problems. However research gap exists in using cultural algorithm for multi-objective optimization of rules. In the current study a multi-objective cultural algorithm is proposed for partial classification. Results of experiments on benchmark data sets reveal good performance.*

## 1. INTRODUCTION

Partial classification popularly known as nugget discovery involves mining rules for a particular class of interest. Classification rules are the most preferred knowledge in the data mining literature since the rules are more comprehensible to the decision maker. The knowledge mined by data mining techniques need to satisfy certain properties specified by the user that are used to evaluate them and render them actionable knowledge. The knowledge discovered is evaluated using these properties known as metrics. The metrics used for evaluating classification rules are both objective and subjective. The objective rule metrics are coverage, support, confidence, precision, recall, sensitivity, specificity, accuracy all of which are used for measuring the accuracy of a rule. While measures like interest, surprise, Jmeasure and comprehensibility, are more user oriented and used to find interesting knowledge to surprise the decision maker. Users prefer to use a combination of these objective and subjective metrics to mine interesting rules. It can be noted that some of these metrics are contradictory. For example it is desired to mine both accurate and interesting rules. The objectives here might be contradictory to each other. Hence discovery of rules can be considered as a multi-objective optimization problem.

Evolutionary algorithms are inspired by evolution occurring in nature and have been effectively used in solving multi-objective problems. They are very efficient in exploring large search spaces. But they are blind search methods which do not retain the knowledge evolved in previous generations. Cultural algorithm (CA) is an evolutionary algorithm which was inspired by social evolution in nature and which uses five knowledge sources to store the evolutionary knowledge and enables efficient exploitation and exploration of the search space. Cultural algorithms use an acceptance function to choose knowledge of previous generation to update the knowledge sources known as the belief space. An influence function is then used to disseminate this stored knowledge in successive generations. This enables cultural algorithms to store the knowledge gained in previous generations and use it immediately in successive generations which in turn enables the evolutionary process to move the search towards better solutions in successive generations enabling faster convergence. Cultural algorithms have been used in solving optimization problems and applied in engineering, function optimization, in rule based systems to reengineer fraud detection and to make archaeological decisions and to study site formation by early inhabitants, to evolve strategies in market problems and to study micro and macro evolution in artificial societies. However research gap exists in applying cultural algorithms for solving the most popular multi-objective rule mining problem. Hence an attempt has been made in the present study to use cultural algorithm to solve the multi-objective rule mining problem, in particular for partial classification. Section 2 defines the problem and related work. Section 3 explains the proposed Multi-objective Cultural Algorithm (MOCA) for nugget discovery. Section 4 explains the experiments and results while section 5 concludes with some future enhancements.

## 2. PROBLEM AND RELATED WORK

### 2.1 The Problem

*Given a data source, the class of interest and multiple objectives namely the rule metrics for optimization specified by the user, the problem is presenting the user with a good set of rules optimized in the user specified metrics.*

### 2.2 Related work

Evolutionary algorithms have been used in the literature to solve multi-objective problems effectively. Although there are a large variety of methods available to produce rule models, there is still a great interest in studying new algorithms capable of producing them since the accuracy of rule models produced by existing rule induction algorithms can still be improved. [1]. Evolutionary algorithms are robust and global search methods that adapt to the environment and can discover interesting knowledge that will be missed by greedy algorithms [2]. Also they allow the user to interactively select interesting properties to be incorporated into the objective function providing the user with a variety of choices [3]. Thus Evolutionary algorithms are very suitable for multi-objective optimization since they allow various objectives to be simultaneously incorporated into the solution. The use of an Evolutionary Multi-objective (EMO) algorithm was proposed to search for Pareto-optimal partial classification initially by Iglesia et al., [3], followed by Ishibuchi and Namba [4]. Iglesia et al. [3], [5], propose the use of multi-objective optimization evolutionary algorithms to allow the user to interactively select a number of interest measures and deliver the best nuggets. Reynolds and Iglesia, [6] describe how the use of modified dominance relations may increase the diversity of rules presented to the user and how clustering techniques may be used to aid in the presentation of the potentially large sets of rules generated. A multi-objective genetic algorithm is applied to the problem of partial classification. An agent-based evolutionary approach is proposed to extract interpretable rule-based knowledge by Wang et al. [7] .While Ishibuchi and Nojima, [8], compare fuzzy rules with interval rules through computational experiments on benchmark data sets using an evolutionary multi-objective rule selection method. Further Ishibuchi et al., [9] as an extension of their previous work explain two approaches to search for Pareto-optimal rules and Pareto-optimal rule sets.

Dehuri and Mall [10] present a multi-objective genetic algorithm for mining highly predictive and comprehensible classification rules from large databases. They have proposed a multi-objective evolutionary algorithm called Improved Niched Pareto genetic algorithm (INPGA) for this purpose. Berlanga et al., [11] and Del Jesus et al., [12] present a multi-objective genetic algorithm for obtaining fuzzy rules for subgroup discovery. The rule induction problem has been considered as a multi-objective combinatorial optimization problem in [13] for finding non frequent and interesting rules. Reynolds and Iglesia [14] use multi-objective genetic programming to induce rules. Baykasoglu and Ozbakir [15] propose a solution technique based on Multi-Expression Programming (MEP) for partial classification. Giusti et al. [16] report a research work that combines evolutionary algorithms and ranking composition methods for multi-objective optimization. In this work candidate solutions are constructed, evaluated and ranked according to their performance in each individual objective. Then rankings are composed into a single ranking which reflects the candidate solutions' ability to solve the multi-objective problem considering all objectives simultaneously. Further Reynolds and Iglesia [17] discuss how to adapt a MO algorithm for the task of partial classification based on a meta-heuristic known as greedy randomized adaptive search procedure (GRASP) while the authors of [18], use the procedure for rule selection. A review of evolutionary algorithms for multi-objective optimization of classification rules is found in [19].

## 2.3  A short survey of Cultural algorithm

Cultural algorithm is an evolutionary algorithm which is mostly applied for optimization problems and which has a set of five Knowledge sources for representing various primitive knowledge's and works on the strategy of survival of the fittest. The agents in the system affect the various Knowledge sources (KS's) and the KS's in turn influence the agents thus directing them towards an optimal solution. Reynolds et al. [20], use cultural algorithm to solve numerical optimization problems to study the micro and macro evolution of the individuals and the system. Cultural algorithm has been used for rule induction as well as rule pruning.  Sternberg and Reynolds [21] use an evolutionary learning approach based on cultural algorithms to learn about the behaviour of a commercial rule-based system for fraud detection. The learned knowledge in the belief space of the cultural algorithm is then used to re-engineer the fraud detection system. Lazar and Reynolds [22] have used genetic algorithms and rough sets for knowledge discovery. Their work combines decision trees and genetic programming for rule induction. Reynolds and Saleem [23] show that the cultural algorithm is more effective than an evolutionary algorithm with only one single-level evolution when they are applied to the problem of finding the new optima in dynamic environments. Reynolds and Peng [24] discuss how the learning of knowledge in the belief space ensures the adaptability of cultural algorithms. Reynolds and Saleem [25] further investigate the contributions of different types of knowledge from the belief space in guiding the search toward the best solutions in both deceptive and non deceptive environments. While Reynolds and Ostrowski [26] have used cultural algorithms to evolve strategies for recessionary markets. Reynolds and Mostafa in [27] propose a Cultural Algorithm Toolkit which allows users to easily configure and visualize the problem solving process of a Cultural Algorithm. The proposed system is applied in solving predator/prey problem in a cones world environment and engineering design.

Reynolds et al., [28] use decision trees to characterize location decisions made by early inhabitants at Monte Alban, a prehistoric urban centre, and have injected these rules into a socially motivated learning system based on cultural algorithms. They have then inferred an emerging social fabric whose networks provide support for certain theories about urban site formation. Ochoa et al. [29] provide a solution of Logistics Service Based on Data Mining in combination with cultural algorithm, and applied it in optimization of distribution of vehicles within a city. A multi-population multi-objective cultural algorithm adopting knowledge migration is proposed by Guo et al., [30] and has been applied to function optimization problems. Reynolds and Liu [31] propose an extension of Cultural Algorithms for Multi-Objective optimization, MOCAT that fully utilizes all of the available categories of knowledge sources and has been applied to function optimization problem. A multi-objective cultural algorithm has been applied to the problem of rule mining by Srinivasan and Ramakrishnan [32]. The proposed MOCA has been used for partial classification in the current study.

## 3. MULTI-OBJECTIVE CULTURAL ALGORITHM

Cultural Algorithm which derives from social structures, and which incorporates evolutionary systems and agents, and uses various knowledge sources for the evolution process better suits the need for solving multi-objective optimization problem and has been used in different domains. CA has three major components: a population space, a belief space, and a protocol that describes how knowledge is exchanged between the first two components. The population space can support any population-based computational model, such as Genetic Algorithms, and Evolutionary Programming [28].

### 3.1. The Belief space

The belief space comprises of the five knowledge sources namely the Normative, Situational, Domain, Topographical and the History KS. For the rule optimization problem the belief space is modified to hold different types of knowledge or Meta data obtained during evolution which is used in successive generations for creating better individuals. Further an additional KS has been added to hold the rules. The agents in the CA have also been given social or cognitive traits which they use in decision making. The following section discusses the different knowledge sources.

#### 3.1.1. Normative KS

Normative Knowledge Source (NKS) contains the attributes and the possible values that the attributes can take. This information is gathered from the training data set. The normative knowledge source is used to store the maximum and minimum values for numeric attributes. For each nominal or discrete attribute, a separate list is maintained that stores the possible values that the attributes can take. The normative KS is updated during train data set creation and used by the agents during mutation.

#### 3.1.2. Situational KS

Situational knowledge source (SKS) consists of the best exemplar found along the evolutionary process. It represents a leader for the other individuals to follow. This way, agents use the example instead of a randomly chosen individual for the recombination. This KS can be updated by storing the best examples at the end of each generation. Agents use these examples for choosing individuals for reproduction. Also the user can specify schema with don't care conditions for certain attributes which can be used by agents for the search of similar/dissimilar individuals to interest the user. In the current study the SKS stores the schema specified by the user as a vector of attribute values.

### 3.1.3. Domain KS

Domain knowledge source (DKS) contains the vector of rule metric values for each rule along with a Rule Identifier (RuleId). Individuals produced by agents are evaluated at the end of each generation and the fitness vector calculated. DKS is updated with these fitness vectors. The fitness vectors in DKS are compared with each other using Pareto optimization strategy to choose elite individuals at the end of each generation. The elite individuals thus chosen are stored in the historical KS.

### 3.1.4. Topographical KS

Diversity maintenance strategy is a characteristic of Multi-objective evolutionary systems for keeping the solutions uniformly distributed in the Pareto optimal set, instead of gathering solutions in a small region only. Restricted mating, where mating is permitted only when the distance between two parents is large enough, is one technique for maintaining diversity of rules [10]. In the proposed system Topographic knowledge is used to store the difference or distance between two rules for the purpose of maintaining diversity of rules. This KS is updated at the end of each generation. The topographical knowledge contains a pair of RuleId's and their dissimilarity measure. Since the attributes are discrete the attribute values in the corresponding positions in the individuals are compared and a value of 0 is assigned to the attributes with similar values and a value of 1 is assigned to dissimilar values. The number of 1's are counted and assigned as the dissimilarity measure for the pair of rules. Hence topographical KS can be used to create novel and interesting rules by choosing pairs of individuals with maximum dissimilarity measure.

### 3.1.5. History KS

History knowledge source (HKS) records in a list, the best individuals along with their RuleId's, and are updated at the end of each generation. Evolutionary algorithms are termed as memory less since they do not retain memory of previous generations. However attempts have been made to retain elite individuals of each generation as a separate elite population to render memory to the evolutionary algorithms. Cultural algorithm renders memory to the evolutionary strategy in a systematic way by using the different knowledge sources. History knowledge can be used to store elite individuals of each generation thus maintaining memory across generations.

### 3.1.6. The rule KS

The cultural algorithm is extended to contain the individuals produced during evolution using the Rule KS (RKS). The representation of the individuals in RKS is similar to that of the HKS. Each entry holds a RuleId and the attribute values as a vector. The RuleId is used as a pointer by the other KS's. RKS is added to CA in order to render memory by maintaining good individuals evolved across generations. New rules are added to RKS at the end of each generation while worst ones are removed.

### 3.1.7. Social Agents

The proposed CA is also extended by adding cognitive traits to the agents. In the original CA the agents are not distinguished but rather considered as having same properties and are used for exploring the solutions. But the proposed CA explicitly distinguishes agents with three traits namely imitator, cautious and risk taker. The agents use this trait in the selection of parents for reproduction using different knowledge sources. Imitators use the situational KS while cautious agents use historical KS for choosing parents for mating. Risk takers are explorers and use any of the different KS's at random. A random integer in the range 0 to size of the corresponding KS is generated and the individual in that particular location in the RKS or SKS or HKS or TKS are chosen and undergo crossover or mutation. If the KS chosen is TKS then the individuals with the maximum dissimilarity measure is chosen from TKS and reproduction operators are applied to the individuals. This enables creation of diverse set of individuals. Cautious agents use only the historical knowledge source while the imitators use the situational knowledge source to create individuals namely the example specified by the user. Reynolds et al., [20], state that agents that use situational and domain knowledge are exploiters while normative and topographical knowledge users are explorers and Historical knowledge users are good trend predictors. The agents can be allowed to change their traits by enabling them to change the type of knowledge source used.

## 3.2. Influence phase

The influence function decides which knowledge sources influence individuals. In the original CA roulette wheel selection based on performance of knowledge sources in the previous generations have been used. In the proposed system selection is left to the agents. In the proposed CAT-CRM the agents use their social trait namely risk taker or imitator or cautious to choose parents for reproduction. Risk takers use knowledge from any of the four knowledge sources namely RKS, HKS, SKS or TKS at random while cautious agents use only the HKS. The imitators use SKS to create individuals using the example specified by the user. NKS which stores the possible attribute values is used by all the agents during the mutation operation. The topographical knowledge source enables creation of a diverse set of rules. DKS stores the values of the user specified metrics of the individuals as a fitness vector and thus is used for comparing individuals using Pareto comparison. The Rule KS is used to store the individuals (both dominators as well as non-dominators thus avoiding losing of good individuals of initial generations) created during evolution. Thus the KS's guide the agents in the evolution process. More social traits can be added to the agents for studying the effect of various traits and knowledge sources on the outcome of the system. The proposed system thus can also be used as a social simulation system or virtual organization which can be used for studying the micro and macro dynamics of real world social systems.

## 3.3. Acceptance phase

The acceptance function determines which individuals and their behaviors can impact the belief space knowledge [20]. Based on selected parameters such as performance, for example, a percentage of the best performers (e.g., top 10%), can be accepted [20]. But since the problem is one of classification rule mining, a threshold value for the rule metrics specified by the user can be used to accept individuals for next generation. Since the current implementation is one of multi-objective optimization, the algorithm produces a set of solutions and the dominators are chosen and stored in HKS using Pareto optimality, while other KS's are updated as explained earlier. The process of agent's selection, reproduction, evaluation and updating of belief space forms a generation. At the end of a generation (iteration), the agents return the individuals created by them along with a fitness vector of rule metric values. The knowledge sources are updated with this new knowledge at the end of each generation and thus evolve along with the agents. The new values in these KS's then influence the population space. Thus the macro evolution takes place.

## 3.4. Evolutionary strategy

Genetic algorithm (GA) is by far the most used evolutionary strategy which is also used in the current study. The various attributes of the GA used are discussed below.

### 3.4.1. Chromosome representation

The chosen data records are converted into fixed size chromosomes and represented as a vector of attribute values. The system uses high level encoding where the attribute values are used as they appear in the data source. This reduces the cost of encoding and decoding individuals for creating rules for large data sets. The relational operators are not included in the genotype as found in most algorithms in the literature. Therefore they are not involved in the reproduction which further minimizes the length of the chromosome and thus the time taken for encoding and/or decoding. This representation also avoids use of different types of reproduction operators for different parts of the chromosome.

In the current study the class attribute is also included in the chromosome during the training phase. During the test phase classes are assigned to individuals as follows: If more than 75% of the values in the antecedent part are equal in the rule created and the test data instance then that class is assigned. If more than one rule covers the test instance then the maximum occurring class label that covers the rule is assigned otherwise maximum occurring class in the data set is assigned.

### 3.4.2. Population initialization

Evolutionary systems work on a population of individuals. Population initialization is an important aspect that decides the overall performance of the algorithm. The initial population is created using various procedures. One procedure is seeding where data instances chosen from training data randomly as in [10] or with the help of the users, are used as initial seeds to fill the population space. Casillas et al., [33] state that the initialization procedure has to guarantee that the initial individuals cover all the input examples from the training data set. To ensure this, the authors of [3] use mutated forms of the default rule as initial solutions where the default rule is the rule in which all limits are maximally spaced and all labels are included. In current study maximum and minimum chromosomes are used as seeds to create initial population. That is, the initialization procedure uses two initial chromosomes as seeds where one chromosome contains the minimum value of all the attributes and the other seed contains the maximum attribute values. These maximum and minimum seeds undergo reproduction and fill the population space.

### 3.4.3. Reproduction operators

The operators used for reproduction are selection, crossover and mutation.

**Selection strategy**

Unlike algorithms found in the EMOO literature, in the proposed system, agents use their social traits in choosing the individuals for reproduction. The agent with the social trait of risk taking chooses rules using any of the knowledge sources at random. The cautious agents choose individuals from historical KS consisting of the elite ones, while imitators use rule schema specified by the user from the situational KS. In this way, knowledge based selection is used rather than random selection. This kind of selection strategy aids in creating not only interesting knowledge but also a diverse set of solutions using the various KS's.

**Crossover**

Multi point uniform crossover is used. Initially two individuals are chosen at random from the population. A random number which is less than the size of the chromosome is chosen and is taken as the number of crossover points say "c". Then "c" random numbers are generated which is again less than the chromosome size and the values at these "c" points are swapped to create two parents.

**Mutation**

Mutation operates on individual values of attributes in the chromosome. A mutation point is chosen similar to that of the crossover point which is a random integer whose value is less than the chromosome size. The value of the attribute at that point is replaced by another value depending upon the type of the value. For nominal and/or discrete attributes the value to be replaced is chosen at random from a list of available values from NKS. If the value is continuous, a random value in a specified range of minimum and maximum values so far encountered is generated and used for reproduction. A list of values for discrete and nominal attributes and lower and upper bound for real valued attributes is stored in the normative knowledge source.

### 3.4.4. Parameters

The parameters that are to be considered and greatly influence the algorithm performance are the crossover rate and the mutation rate. Also the population size and the number of generations or the termination condition are parameters of importance.

**Crossover rate**

There are a variety of issues that have been discussed in the literature regarding the crossover rate. Experiments have been carried out using varying crossover rates ranging from 0% to 100% but hardly any optimum value has been reported.

**Mutation rate**

Mutation rate is the rate at which mutation occurs in a generation. A low mutation rate of 1% and a high mutation rate of 20% can be found in the literature. However there are discussions about varying the mutation rate as the algorithm proceeds.

**Population size**

Population sizes ranging from a few dozens to hundreds have been reported. Population size can be varied depending upon the size of the data source.

**Stopping criteria**

Stopping criteria can be set to a certain number of generations, or it can be set by the user, where the user can stop the algorithm at a point where a satisfactory set of rules have been obtained. Another condition which can be used as stopping criteria is coverage of all the records in the train data set. The algorithm stops when all the records in the train data set have been covered by at least a single rule known as the sequential covering approach. In the current study number of generations specified by the user is used as the stopping criteria. Table 1 summarizes the parameter settings used in the experiments.

Table 1 Parameter settings

| Parameters | Values |
|---|---|
| Crossover rate | 80% |
| Mutation rate | 20% |
| No. of generations | 25 |
| Stopping criteria | No. of generations |
| Initialization process | Seeding |
| Optimization strategy | Pareto optimality |
| Optimization Metrics | Support and Confidence |

## 3.5. Optimization strategy/Fitness evaluation

The optimization or multi-objective optimization strategy forms the acceptance phase of the cultural algorithm. The ultimate objective of multi-objective algorithms is to guide the user's decision making, through the provision of a set of solutions that have differing trade-offs between the various objectives  [Reynolds A. P. and de la Iglesia B., [14], and thus the user must be involved in the process of discovering rules . Therefore in the proposed system the user is allowed to control the system by specifying most of the attributes of the system including the rule metrics (objectives), the rule schema, and other parameters as discussed earlier. The user can choose any combination of metrics including coverage, support, confidence, interest, surprise, precision, recall/sensitivity, specificity and a difference measure that stores the difference between the rule and the user specified schema. Coverage and Confidence have been used in the current study. Pareto optimality and ranking composition methods are the frequently used optimization strategies. In the current study Pareto optimality has been used as the optimization strategy to select elite individuals.

Pareto optimality is an optimization strategy that uses comparison of the metrics represented as a vector. An individual "A" is said to be better than another individual "B" if "A" is better than "B" in all the metric values or equal to "B" in all but one metric and better at least in one metric value. This is enabled by the use of Domain KS which stores the rule metrics as fitness vectors. The entries in the DKS are compared with each other and the best performers in all the metrics are returned as dominators. The dominators form the Pareto front found in the Historical KS at the end of the algorithm execution.

## 4. EXPERIMENTS, RESULTS AND DISCUSSION

### 4.1  Experiments

Experiments were conducted on two data sets in the medicinal domain from the UCI machine learning repository [34]. The Ljubljana breast cancer (LJB) data set and the Wisconsin breast cancer (WBC) data set are used in the experiments. Table 2 summarizes the data set information. The population size was taken as 300 for LJB data set and 500 for WBC data set.

Table 2 Data set information

| Data sets | Attributes | Patterns | Classes |
|---|---|---|---|
| Ljubljana Breast cancer (LJB) | 9 | 277 | 2 |
| Wisconsin Breast cancer (WBC) | 9 | 683 | 2 |

## 4.2 Results

Table 3 gives the comparative performance of the two types of classification, parallel and partial for the Ljubljana breast cancer data set. The table summarizes the average over ten runs of the number of unique rules created by the algorithm as in RKS, the number of dominators as in HKS, the time taken by the algorithm and the accuracy on the test data for the LJB data set. Best values are shown in bold face.

Table 3 Comparative performance summary of parallel and partial classification for LJB data

| Type of classification | Measures | RKS | HKS | Time (seconds) | Accuracy% |
|---|---|---|---|---|---|
| Parallel | Average | 126.9 | 10.1 | 3.08 | 90.86 |
| | Stdev | 12.66 | 5.63 | 0.85 | 1.70 |
| | Min | 107 | 3 | 2.43 | 86.02 |
| | Max | 148 | 19 | 4.52 | 91.4 |
| Partial | Average | 111.75 | 8.05 | 2.18 | **98.20** |
| | Stdev | 6.74 | 4.50 | 0.83 | 2.61 |
| | Min | 102 | 2 | 3.43 | 93.10 |
| | Max | 123 | 17 | 3.70 | **100** |

Table 4 gives the comparative performance of the two types of classification, parallel and partial for the Wisconsin breast cancer data set. The table summarizes the average over ten runs of the number of unique rules created by the algorithm as in RKS, the number of dominators as in HKS, the time taken by the algorithm and the accuracy on the test data for WBC data set. Best values are shown in bold face.

Figure 3 gives a sample set of dominators for the LJB data set. Fig 4 gives a sample set of dominators in support and confidence for WBC data set. The confidence values for the WBC data set were considered without rounding for drawing the graph since the values obtained had very small difference and tended to 1 in almost all the cases.

Table 4 Comparative performance summary of parallel and partial classification for WBC data

| Type of classification | Measures | RKS | HKS | Time (seconds) | Accuracy% |
|---|---|---|---|---|---|
| Parallel | Average | 319.10 | 23.7 | 12.97 | 95.40 |
| | Stdev | 11.01 | 11.14 | 6.27 | 1.31 |
| | Min | 303 | 6 | 6.42 | 93.86 |
| | Max | 335 | 39 | 25.51 | 97.37 |
| Partial | Average | 192.25 | **10.7** | **5.12** | **99.33** |
| | Stdev | 12.02 | 5.52 | 1.20 | **0.61** |
| | Min | 149 | 3 | 5.33 | **97.42** |
| | Max | 243 | 25 | 8.59 | **100** |

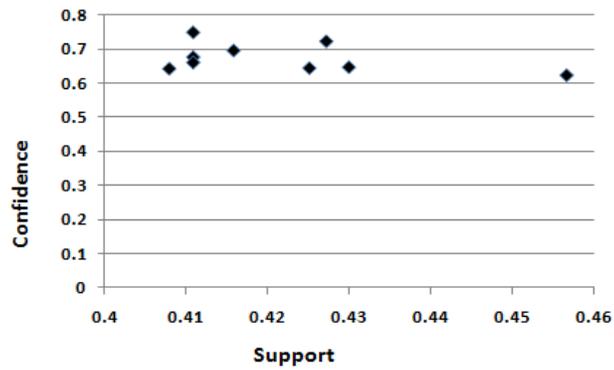

Figure 1 Sample set of dominators for LJB data set

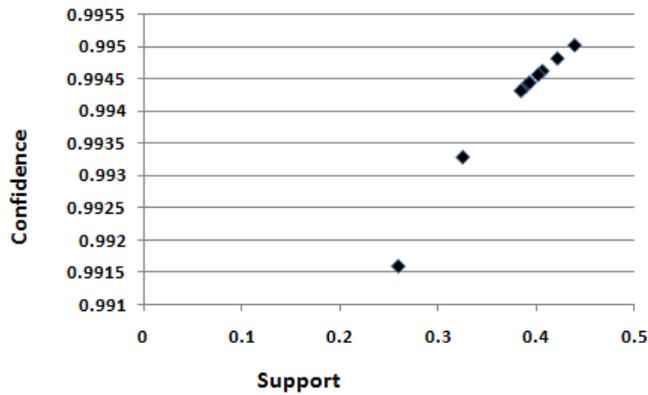

Figure 2 Sample set of dominators for WBC data set

## 4.3 Discussion

The results of the two types of classifications, partial and parallel are summarized in Tables 3 and 4 for the two data sets. It can be noted that in both the data sets the accuracy is higher in the case of partial classification. From Table 3 for the LJB data set it can be noted that the average accuracy is 90.86% when parallel classification was used for all the classes taken simultaneously whereas when using partial classification the average accuracy is 98.20% which is significantly large. Also the standard deviation is lesser in the case of partial classification than parallel. It can also be noted that both the number of unique rules created and the number of dominators are less in number. This is also observed in the WBC data set as found in Table 4. This has an important implication in that, the number of rules in RKS constitute the highest time taken for comparison of the rules with the data instances. Thus less number of rules in RKS implies reduction in time taken by the algorithm to complete. Also less number of rules in HKS implies higher comprehensibility by presenting the user with a compact set of rules. More number of rules during parallel rule induction can be attributed to the fact that the chromosome representation also includes the class attribute. The number of rules in RKS can be reduced if the class attribute is not included in the chromosome representation but rather assigned at runtime. However this will lead to more comparisons and calculations leading to more time for completion, which is the reason for using this type of chromosome representation in the current implementation. There is no significant difference in the time taken for partial and parallel classification for the LJB data set. The average time taken is around 2 to 3 seconds with a maximum of 4.52 for parallel classification. However when considering the WBC data set, the average time taken for parallel classification is 25.51 seconds as compared to 8.59 seconds for partial classification. The more deviation in time can be attributed to the size of the WBC data set with 683 instances. The average accuracy for parallel classification is less with 95.40% as compared to partial classification with 99.33%. It can be observed that, for both the data sets the maximum accuracy of 100% is obtained during partial classification. However for the WBC data set the maximum accuracy achieved when using parallel classification is 97.37% whereas for LJB data the maximum is only 91.40%. Figure 1 and Figure 2 gives sample set of dominators for the LJB and WBC data sets respectively. It can be observed from both figures that the algorithm is good in mining accurate rules with high support and confidence.

Table 5 Comparative Performance summary with other algorithms for the WBC data set

| Reference | Accuracy% | No. of dominators | Time |
|-----------|-----------|-------------------|------|
| [8] | 99.79(Train data) 64.98 (Test data) (10 fold CV) | NS | Min: 4 minutes Max: 170.6 minutes |
| [9] | 99.7 (Train data) 95.6 (Test data) | Min: 10.4 Max: 16.7 (Rule sets) | NS |
| [33] | 96.24 (Best) 94.81(Median) | NS | NS |
| MOCA | 100 (Best) 99.33 (Average) | 10.7 (Rules) | 5.12 seconds |

NS - Not specified

Table 5 gives a comparison of performance of MOCA with other recent multi-objective algorithms for the WBC data set since it is the most frequently used data set found in the literature. Again it can be noted that the proposed MOCA performs better than most of the algorithms both in terms of accuracy and in producing a compact set of rules. Moreover the algorithms found in the literature are rule selection methods while the proposed MOCA both induces and selects rules immediately at the end of each generation. However the three algorithms used for comparison with MOCA are ones which return Pittsburgh style classifiers in the form of rule sets or rules in conjunctive normal form while MOCA returns Michigan style set of rules. Hence the comparison in terms of number of rules returned by the algorithm and the time taken for the algorithm to complete is a bit complex. Moreover the algorithm in [8] and [9] are rule selection algorithms where the candidate rules fed to the algorithm for rule selection are generated by another rule mining algorithm. Whereas MOCA generates the candidate rules, evaluates them immediately and selects the best rules at the end of each generation. Thus in successive generations the algorithm uses better individuals for reproduction and thus converges in less number of generations. The proposed MOCA also performs better than the other algorithms in classifying unknown data instances with a maximum accuracy of 100% and an average accuracy of 99.33%. This illustrates the influence of the added evolutionary knowledge in improving the algorithm performance.

## 5. CONCLUSION

A multi-objective cultural algorithm was proposed for nugget discovery taking classification rule mining as a multi-objective optimization problem. The algorithm performs well in terms of accuracy as well as time taken to converge in comparison to other recent multi-objective algorithms. The proposed cultural algorithm enables adding the evolutionary knowledge obtained in each generation to be incorporated into successive generations through the use of the different knowledge sources. This enables faster convergence towards better solutions. However the algorithm needs testing on larger data sets with more data instances. Moreover in an earlier version the number of generations taken to converge was more. But when a procedure to ensure that the initial population consisted of rules which are consistent with the training data was added by using the RKS and DKS the number of generations to converge reduced. Therefore addition of meta heuristics along with the meta data in the knowledge sources would further increase the performance of the algorithm. This gives new avenues for future research. Moreover most of the parameters of the algorithm can be controlled by the user and hence the system can be used as a toolkit for experimenting with classification rule mining.

## Authors

**Sivakumar Ramakrishnan** is Reader and Head of the Research Department of Computer Science in AVVM Sri Pushpam College, Tamil Nadu, India since 1987. His research interests include Data mining, Human Computer Interaction and Bio-informatics. He has published a number of papers in National and International Journals. He received his PhD in Computer Science from Barathidasan University, India in the year 2005.

**Sujatha Srinivasan** received her Master's degree in Mathematics in 1993 and Master's degree in Computer Applications in 2000. She received her Master of Philosophy in Computer Science in 2004. She is Assistant Professor in the PG and Research department of Computer Science in Cauvery College for women, Tamil Nadu, India for the past ten years. She is a Research scholar in AVVM Sri Pushpam College, India. Her research interests include Simulation modeling, Data mining, Human Computer Interaction and Evolutionary computing. She has published papers in International Journals and presented papers in International Conferences.